\RequirePackage{fix-cm}
\documentclass[twocolumn]{svjour3}          
\smartqed  

\newtheorem{guess}{Hypothesis}
\newtheorem{obs}{Observation}

\usepackage{graphicx}

\usepackage{hyperref}
\hypersetup{colorlinks,linktocpage,plainpages=false, bookmarksopen = true}
\usepackage{multirow}
\usepackage{amsmath}
\usepackage{amssymb}
\usepackage{wrapfig}
\usepackage{soul}

\newcommand*{\Perm}[2]{{}^{#1}\!\mathbf{P}_{\!#2}}

\newcommand \myeq{\stackrel{\normalfont \mbox{\tiny{L'H\^{o}pital's}}}{=}}

\usepackage[ruled,linesnumbered]{algorithm2e}
\journalname{arXiv}
\begin{document}
\pagenumbering{arabic}
\sloppy

\title{The training accuracy of two-layer neural networks: its estimation and understanding using random datasets 
}

\titlerunning{The training accuracy of two-layer neural networks: its estimation and...}        

\author{Shuyue Guan         \and
        Murray Loew 
}


\institute{Shuyue Guan \at
              ORCID: \href{https://orcid.org/0000-0002-3779-9368}{0000-0002-3779-9368} \\
              \email{frankshuyueguan@gwu.edu}           
           \and
           Murray Loew \at
              \email{loew@gwu.edu} \\
              \\
           Department of Biomedical Engineering, \\
           George Washington University, Washington DC, USA
}

\date{}

\maketitle

\begin{abstract}
Although the neural network (NN) technique plays an important role in machine learning, understanding the mechanism of NN models and the \textit{transparency} of deep learning still require more basic research. In this study we propose a novel theory based on space partitioning to estimate the approximate training accuracy for two-layer neural networks on random datasets without training. There appear to be no other studies that have proposed a method to estimate training accuracy without using input data and/or trained models. Our method estimates the training accuracy for two-layer fully-connected neural networks on two-class random datasets using only three arguments: the dimensionality of inputs ($d$), the number of inputs ($N$), and the number of neurons in the hidden layer ($L$). We have verified our method using real training accuracies in our experiments. The results indicate that the method will work for any dimension, and the proposed theory could extend also to estimate deeper NN models. The main purpose of this paper is to understand the mechanism of NN models by the approach of estimating training accuracy but not to analyze their generalization nor their performance in real-world applications. This study may provide a starting point for a new way for researchers to make progress on the difficult problem of understanding deep learning.
\end{abstract}

\section{Introduction}
\label{sec:1}
In recent years, the neural network (deep learning) technique has played a more and more important role in applications of machine learning. To comprehensively understand the mechanisms of neural network (NN) models and to explain their output results, however, still require more basic research \cite{roscher_explainable_2020}. To understand the mechanisms of NN models, that is, the \textit{transparency} of deep learning, there are mainly three ways: the training process \cite{du_gradient}, generalizability \cite{liu_understanding_2020}, and loss or accuracy prediction \cite{arora_fine-grained_2019}.

\begin{figure}[h] 
    \includegraphics[width=0.45\textwidth]{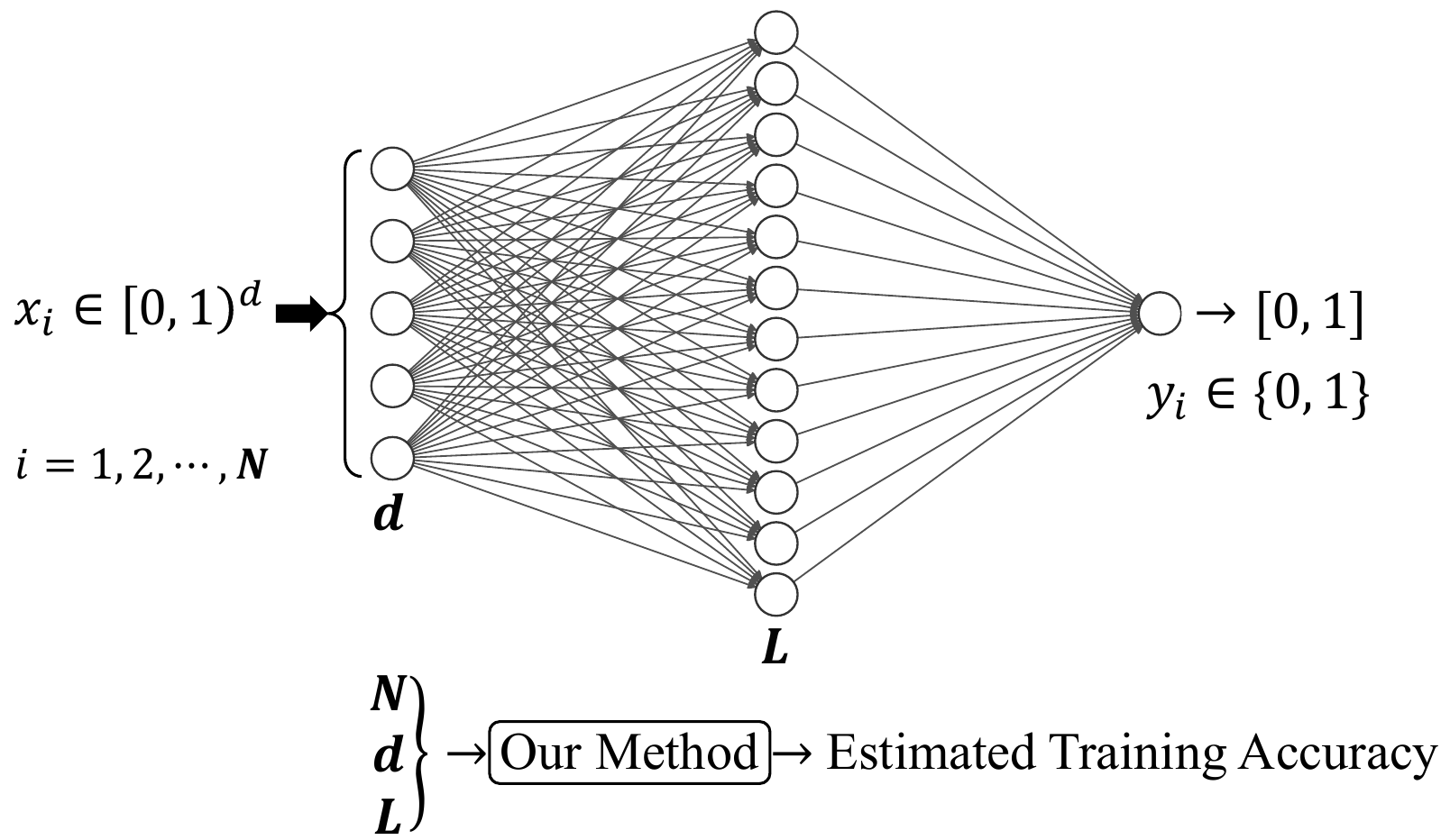}
    \caption{An example of the two-layer FCNN with $d-L-1$ architecture. This FCNN is used to classify $N$ random vectors in $R^d$ belonging to two classes. Detailed settings are stated before in Sec.~\ref{sec:prelim}. The training accuracy of this classification can be estimated by our proposed method, without applying any training process. The detailed \hyperref[algorithm]{Algorithm} of our method is shown in Sec.~\ref{sec:sum_alg}.} \label{fig:two_layer_nn}
\end{figure}

In this study, we create a novel theory from scratch to estimate the training accuracy for two-layer neural networks applied to random datasets. Figure~\ref{fig:two_layer_nn} demonstrates the mentioned two-layer neural network and summarizes the processes to estimate its training accuracy using the proposed method. Its main idea is based on the regions of linearity represented by NN models \cite{pascanu_number_2014}, which derives from common insights of the Perceptron.

This study may raise other questions and offer the starting point of a new way for future researchers to make progress in the understanding of deep learning. Thus, we begin from a simple condition of two-layer NN models, and we discuss the use for multi-layer networks in Sec.~\ref{sec:4.2.2} as future works.
This paper has two main contributions: 
\begin{itemize}
    \item We propose a novel theory to understand the mechanisms of two-layer FCNN models.
    \item By applying that theory, we estimate the training accuracy for two-layer FCNN on random datasets.
\end{itemize}
More discussion about our contributions are in Sec.~\ref{sec:4.1}.

\subsection{Preliminaries}
\label{sec:prelim}
Specifically, the studied subjects are:
\begin{itemize}
  \item \textbf{Classifier model}: the two-layer fully-connected neural networks (FCNN) with $d-L-1$ architecture, in which the length of input vectors ($\in R^d$) is d, the hidden layer has $L$ neurons (with ReLU activation), and the output layer has one neuron, using the Sigmoid activation function. This FCNN is for two-class classifications and outputs of the FCNN are values in [0,1]
  \item \textbf{Dataset}: $N$ random (uniformly distributed) vectors in $R^d$ belonging to two classes with labels ‘0’ and ‘1’, and the number of samples for each class is the same. We consider that the uniformly-distributed dataset is an extreme situation for classification; its predicted accuracy is thus the lower-bound for all other situations.
  \item \textbf{Metrics}: training accuracy.
\end{itemize}

We focus on the training accuracy instead of the test accuracy because the main idea is to understand the mechanism of NN models by the approach of estimating training accuracy, but not to analyze their performances.
The paradigm we use to study the FCNN is the following:
\begin{enumerate}
    \item We find a simplified system to examine.
    \item We create a theory based on the Hypotheses~\ref{hyp:1} and \ref{hyp:2} to predict or estimate outputs of the system.
    \item For the most important step, we perform experiments to test the proposed theory by comparing actual outputs of the system with predicted outputs. If the predictions are close to the real results, we could accept the theory, or update it to make predictions/estimates more accurate by these heuristic observations (empirical corrections). Otherwise, we abandon this theory and seek another one. 
\end{enumerate}

\subsection{Related work}
To the best of our knowledge, only a few studies have discussed the prediction/estimation of the training accuracy of NN models. None of them, however, estimates training accuracy without using input data and/or trained models as does our method.

The overall setting and backgrounds of the studies of \textit{over-parameterized} two-layer NNs~\cite{du_gradient,arora_fine-grained_2019} are similar to ours. But the main difference
is that they do not estimate the value of training accuracy. One study~\cite{du_gradient} mainly shows that the zero training loss on deep over-parametrized networks can be obtained by using gradient descent. Another study analyzes the \textit{generalization bound}~\cite{arora_fine-grained_2019} between training and test performance (\textit{i.e.}, generalization gap). There are other studies~\cite{DBLP:conf/iclr/JiangKMB19,DBLP:journals/corr/abs-1906-01550} to investigate the prediction of the generalization gap of neural networks. We do not further discuss the generalization gap because we focus on only the estimation of training accuracy, ignoring the test accuracy.

Unlike our proposed method that does not need to use input data nor to apply any training process, in recent works related to the accuracy estimation for neural networks~\cite{yamada_weight_2016,Gao_2021_CVPR,chen_practical_2020,DBLP:conf/ijcai/ShaoYR20}, the accuracy prediction methods require pre-trained NN models or weights from the pre-trained NN models. Through our method, to estimate the training accuracy for two-layer FCNN on random datasets (two classes) requires only three arguments: the dimensionality of inputs ($d$), the number of inputs ($N$), and the number of neurons in the hidden layer ($L$).
The \textit{Peephole}~\cite{deng_peephole} and \textit{TAP}~\cite{istrate_tapas_2019} techniques apply Long Short Term Memory (LSTM)-based frameworks to predict a NN model’s performance before training the original NN model. However, the frameworks themselves still must be trained by the input data before making predictions.

\section{The hidden layer: space partitioning}
\label{sec:2}
In general, the output of the $k$-th neuron in the first hidden layer is:
\[
s_k(x)=\sigma(w_k\cdot x+b_k),
\]
where input $x\in R^d$; parameter $w_k$ is the input weight of the $k$-th neuron and its bias is $b_k$. We define $\sigma(\cdot)$ as the ReLU activation function, defined as:
\[
\sigma(x)=\max\{0,x\}.
\]

The neuron can be considered as a hyperplane: $w_k\cdot x+b_k=0$ that divides the input space $R^d$ into two partitions \cite{pascanu_number_2014}. If the input $x$ is in one (lower) partition or  on the hyperplane, then $w_k\cdot x+b_k\leq 0$ and then their output $s_k(x)=0$. If $x$ is in the other (upper) partition, its output $s_k(x)>0$. Specifically, the distance from $x$ to the hyperplane is:
\[
d_k(x)=\frac{|w_k\cdot x+b_k|}{\|w_k\|}
\]
If $w_k\cdot x+b_k>0$,
\[
s_k(x)=\sigma(w_k\cdot x+b_k)=|w_k\cdot x+b_k|=d_k(x)\|w_k\|.
\]

For a given input data point, $L$ neurons assign it a unique code: \{$s_1$, $s_2$, $\cdots$, $s_L$\}; some values in the code could be zero. $L$ neurons divide the input space into many partitions, input data in the same partition will have codes that are more similar because of having the same zero positions. Conversely, it is obvious that the codes of data in different partitions have different zero positions, and the differences (the Hamming distances) of these codes are greater.  It is apparent, therefore, that the case of input data separated into different partitions is favorable for classification.

\subsection{Complete separation}
\label{sec:2.1}
Given $L$ neurons that divide the \textbf{input space} into $S$ partitions, we hypothesize that: 
\begin{guess} \label{hyp:1}
In anticipation of classification, but without yet assigning labels, we identify the best case as the separation of all $N$ input data into different partitions (\textbf{complete separation}). 
\end{guess}
\begin{remark}
For most real classification problems (\textit{i.e.}, in which labels have been assigned), complete separation of all data points is a very strong assumption because adjacent same-class samples assigned to the same partition is a looser condition and will not affect the classification performance. Because \textbf{our partitions by complete separation are unlabeled}, the principle of discriminant analysis (which aims to minimize the
within-class separations and maximize the between-class separations) is not applicable.
Finally, adjacent samples assigned to different partitions could have the same label and thus define a within-class separation. The partitions we mentioned are thus not necessarily the final decision regions for classification; those will be determined when labels are assigned.
\end{remark}
Under this hypothesis, for complete separation, each partition contains at most one data point after space partitioning. Since the position of data points and hyper-planes can be considered (uniformly distributed) random (because the methods to find hyper-planes themselves contain randomness, \textit{e.g.}, the \textit{Stochastic Gradient Descent}~(SGD)~\cite{sutskever2013importance}, during the training), the probability of complete separation ($P_c$) is:

\begin{equation} \label{eq:1}
P_c=\frac{\Perm{S}{N}}{S^N} = \frac{S!}{\left(S-N\right)!S^N}
\end{equation}
In other words, $P_c$ is the probability that each partition contains at most one data point after randomly assigning $N$ data points to $S$ partitions. By Stirling's approximation,
\[
P_c=\frac{S!}{\left(S-N\right)!S^N}\approx\frac{\sqrt{2\pi S}\left(\frac{S}{e}\right)^S}{\sqrt{2\pi\left(S-N\right)}\left(\frac{S-N}{e}\right)^{S-N}S^N}
\]
\begin{equation} \label{eq:2}
P_c=\left(\frac{1}{e}\right)^N\left(\frac{S}{S-N}\right)^{S-N+0.5}
\end{equation}
Let $S=bN^a$ ($a$ and $b$ are two coefficients to be determined) and for a large $N\rightarrow\infty$, by Eq.~(\ref{eq:2}), the limitation of complete separation probability is:
\begin{equation} \label{eq:3}
\lim_{N\rightarrow\infty}P_c=\lim_{N\rightarrow\infty}\left(\frac{1}{e}\right)^N{\left(\frac{bN^a}{bN^a-N}\right)^{bN^a-N+0.5}}
\end{equation}
$S>0$ requires $b>0$; and for complete separation, $\forall N:\ S\geq N$ (Pigeonhole principle) requires $a\geq1$. By simplifying the limit in Eq.~(\ref{eq:3}), we have\footnote{A derivation of this simplification is in the \hyperref[sec:appx]{Appendix}.}:
\begin{equation} \label{eq:4}
\begin{array}{l p{1em} r r}
\displaystyle \lim_{N\rightarrow\infty}P_c=\lim_{N\rightarrow\infty}e^{-\frac{(a-1)N^{2-a}}{ab}} & & \mbox{when} & a>1 \\
& & &\\
\displaystyle \lim_{N\rightarrow\infty}P_c=0 & & \mbox{when} & a=1
\end{array}
\end{equation}
Eq.~(\ref{eq:4}) shows that for large $N$, the probability of complete separation is \textbf{nearly zero} when $1\le a<2$, and \textbf{close to one} when $a>2$. Only for $a=2$ (\textit{i.e.}, $S=bN^2$) is the probability controlled by the coefficient $b$:
\begin{equation} \label{eq:5}
\displaystyle \lim_{N\rightarrow\infty}P_c=e^{-\left(\frac{1}{2b}\right)}
\end{equation}
Although complete separation holds $\left(\displaystyle \lim_{N\rightarrow\infty}P_c=1\right)$ for $a>2$, there is no need to incur the exponential growth of $S$ with $a$ when compared to the linear growth with $b$. And a high probability of complete separation does not require even a large $b$. For example, when $a=2 \mbox{ and } b=10$ the $\displaystyle \lim_{N\rightarrow\infty}P_c\approx0.95$. Therefore, we let $S=bN^2$ throughout this study.

\subsection{Incomplete separation}
\label{sec:2.2}
To increase the $b$ in Eq.~(\ref{eq:5}) can improve the probability of complete separation. Alternatively, decreased training accuracy could improve the probability of an \textbf{incomplete separation}; that is, some partitions have more than one data point after space partitioning. We define the separation ratio $\gamma\ \left(0\le\gamma\le1\right)$ for $N$ input data, which means at least $\gamma N$ data points have been completely separated (at least $\gamma N$ partitions contain only one data point). According to  Eq.~(\ref{eq:1}), the  probability of such incomplete separation ($P_{inc}$) is:
\begin{equation} \label{eq:6}
P_{inc}=\frac{\Perm{S}{\gamma N}\cdot {\left(S-\gamma N\right)^{\left(N-\gamma N\right)}}}{S^N} = \frac{S!\left(S-\gamma N\right)^{\left(N-\gamma N\right)}}{\left(S-\gamma N\right)!S^N}
\end{equation}
In other words, $P_{inc}$ is the probability that at least $\gamma N$ partitions contain only one data point after randomly assigning $N$ data points to $S$ partitions. When $\gamma=1$, $P_{inc}=P_c$, \textit{i.e.}, it becomes the complete separation, and when $\gamma=0$, $P_{inc}=1$. We apply Stirling's approximation and let $S=bN^2$, $N\rightarrow\infty$, similar to  Eq.~(\ref{eq:5}), we have:
\begin{equation} \label{eq:7}
\lim_{N\rightarrow\infty}P_{inc}=e^\frac{\gamma\left(\gamma-2\right)}{2b}
\end{equation}

\subsection{Expectation of separation ratio}
\label{sec:2.3}
In fact, Eq.~(\ref{eq:7}) shows the probability that \textbf{at least} $\gamma N$ data points (when $N$ is large enough) have been completely separated, which implies:
\begin{multline*}
P_{inc}\left(x\geq\gamma\right)=e^\frac{\gamma\left(\gamma-2\right)}{2b}\Rightarrow \\ P_{inc}\left(x=\gamma\right)=\frac{dP_{inc}\left(x<\gamma\right)}{d\gamma}=\frac{d\left(1-P_{inc}\left(x\geq\gamma\right)\right)}{d\gamma} \\ =
\frac{d\left(1-e^\frac{\gamma\left(\gamma-2\right)}{2b}\right)}{d\gamma}=\frac{1-\gamma}{b}e^\frac{\gamma\left(\gamma-2\right)}{2b}=P_{inc}\left(\gamma\right)
\end{multline*}
We notice that the equation $P_{inc}\left(\gamma\right)$ does not include the probability of complete separation $P_c$ because $P_{inc}\left(1\right)=0$. Hence, $P_{inc}\left(1\right)$ is replaced by $P_c$ and the comprehensive probability for the separation ratio $\gamma$ is:
\begin{equation} \label{eq:8}
P(\gamma)=\left\{
\begin{array}{l p{1em} c}
P_c=e^{-\left(\frac{1}{2b}\right)}& & \gamma=1 \\
\frac{1-\gamma}{b}e^\frac{\gamma\left(\gamma-2\right)}{2b}& & 0\le\gamma<1
\end{array}\right.
\end{equation}
Since Eq.~(\ref{eq:8}) is a function of probability, we could verify it by observing:
\begin{multline*}
\int_{0}^{1}P\left(\gamma\right)d\gamma=P_c+\int_{0}^{1}{\frac{1-\gamma}{b}e^\frac{\gamma\left(\gamma-2\right)}{2b}d\gamma}\\ =e^{-\left(\frac{1}{2b}\right)}+\left(1-e^{-\left(\frac{1}{2b}\right)}\right)=1
\end{multline*}
We compute the expectation of the separation ratio $\gamma$:
\begin{multline*}
E\left[\gamma\right]=\int_{0}^{1}{\gamma\cdot P\left(\gamma\right)d\gamma}\\ =1\cdot P_c+\int_{0}^{1}{\gamma\cdot\frac{1-\gamma}{b}e^\frac{\gamma\left(\gamma-2\right)}{2b}d\gamma}\Rightarrow
\end{multline*}
\begin{equation} \label{eq:9}
E\left[\gamma\right]=\frac{\sqrt{2\pi b}}{2}\mbox{erfi}{\left(\frac{1}{\sqrt{2b}}\right)}e^{-\left(\frac{1}{2b}\right)}
\end{equation}
where $\mbox{erfi}(x)$ is the imaginary error function:
\[
\mbox{erfi}(x)=\frac{2}{\sqrt\pi}\sum_{n=0}^{\infty}\frac{x^{2n+1}}{n!\left(2n+1\right)}
\]

\subsection{Expectation of training accuracy}
\label{sec:2.4}
Based on the hypothesis, a high separation ratio helps to obtain a high training accuracy, but it is not sufficient because the training accuracy also depends on the separating capacity of the second (output) layer. Nevertheless, we initially ignore this fact and reinforce our Hypothesis~\ref{hyp:1}.
\begin{guess} \label{hyp:2}
The separation ratio directly determines the training accuracy.
\end{guess}

Then, we will add \textbf{empirical corrections} to our theory to allow it to match the real situations. In the case of incomplete separation, 1) all completely-separated data points can be predicted correctly, and 2) the other data points have a 50\% chance to be predicted correctly (equivalent to a random guess, since the number of samples for each class is the same) because each partition will be ultimately assigned one label. Specifically, if $ \gamma N$ data points have been completely separated, the training accuracy $ \alpha$ (based on our hypothesis) is:
\[
\alpha=\frac{\gamma N+0.5\left(1-\gamma\right)N}{N}=\frac{1+\gamma}{2}
\]
To take the expectation on both sides, we have:
\begin{equation} \label{eq:10}
E\left[\alpha\right]=\frac{1+E\left[\gamma\right]}{2}
\end{equation}
Eq.~(\ref{eq:10}) shows the expectation relationship between the separation ratio and training accuracy. After replacing $E\left[\gamma\right]$ in Eq.~(\ref{eq:10}) with Eq.~(\ref{eq:9}), we obtain the formula to compute the \textbf{expectation of training accuracy}:
\begin{equation} \label{eq:11}
\boxed{
E\left[\alpha\right]=\frac{1}{2}+\frac{\sqrt{2\pi b}}{4}\mbox{erfi} {\left(\frac{1}{\sqrt{2b}}\right)}e^{-\left(\frac{1}{2b}\right)}
}
\end{equation}

To compute the expectation of training accuracy by Eq.~(\ref{eq:11}), we must calculate the value of $b$. The expectation of training accuracy is a \textbf{monotonically increasing function of $b$ on its domain $(0,\infty)$ and its range is $(0.5,1)$}. Since the coefficient $b$ is very important in estimation of the training accuracy, it is also called the \textbf{ensemble index} for training accuracy.  This leads to the following theorem: 

\begin{theorem} \label{thm:1}
The expectation of training accuracy for a $d-L-1$ architecture FCNN is determined by Eq.~(\ref{eq:11}) with the ensemble index $b$.
\end{theorem}
In the input space $R^d$, $L$ hyperplanes (neurons) divide the space into $S$ partitions. By the Space Partitioning Theory~\cite{winder_partitions_1966}, the maximum number of partitions is:
\begin{equation} \label{eq:12}
S=\sum_{i=0}^{d}{L \choose i}
\end{equation}
Since:
\[
\sum_{i=0}^{d}{L \choose i}=O\left(\frac{L^d}{d!}\right)
\]
We let:
\begin{equation} \label{eq:13}
S=\frac{L^d}{d!}
\end{equation}

\begin{figure}
    \includegraphics[width=0.45\textwidth]{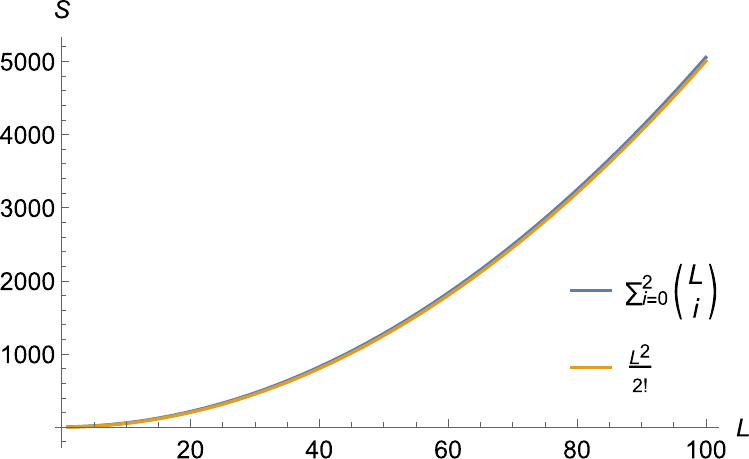}
    \caption{Maximum number of partitions in 2-D} \label{fig:1}
\end{figure}

Figure~\ref{fig:1} shows that the partition numbers calculated from Eqs.~(\ref{eq:12}) and (\ref{eq:13}) are very close in 2-D. In high dimensions, both theory and experiments show that Eq.~(\ref{eq:13}) is still an \textbf{asymptotic upper-bound} of Eq.~(\ref{eq:12}). By our agreement in Eq.~(\ref{eq:5}), which $S=bN^2$; we have:
\begin{equation} \label{eq:14}
b=\frac{L^d}{{d!N}^2}
\end{equation}

Now, we have introduced our main theory that could estimate the training accuracy for a $d-L-1$ structure FCNN and two classes of $N$ random (uniformly distributed) data points by using Eq.~(\ref{eq:14}) and (\ref{eq:11}). For example, let a dataset have 200 two-class random data samples in $R^3$ (100 samples for each class) and let it be used to train a $3-200-1$ FCNN. In this case,
\[
b=\frac{{200}^3}{3!\cdot{200}^2}\approx33.33.
\]
Substituting $b=33.33$ into Eq.~(\ref{eq:11}) yields $E\left[\alpha\right]\approx0.995$, \textit{i.e.}, the expectation of training accuracy for this case is about 99.5\%.

\section{Empirical corrections}
\label{sec:3}
The empirical correction uses results from real experiments to update/correct the theoretical model we have proposed above. The correction is necessary because our hypothesis ignores the separating capacity of the second (output) layer and because the maximum number of partitions is not guaranteed for all situations; \textit{e.g.}, for a large $L$, the real partition number may be much smaller than $S$ in Eq.~(\ref{eq:12}). 

In experiments, we train a $d-L-1$ structure FCNN by two-class $N$ random (uniformly distributed) data points in $[0,\ 1)^d$ with labels ‘0’ and ‘1’ (and the number of  samples for each class is equal). The training process ends when the training accuracy converges (loss change is smaller than ${10}^{-4}$ in 1000 epochs). For each $\left\{d,\ N,\ L\right\}$, the training repeats several times from scratch, and the recorded training accuracy is the average.

\subsection{Two dimensions; empirical correction procedure}
\label{sec:3.1}
In 2-D, by Eq.~(\ref{eq:14}), we have:
\[
b=\frac{1}{2}\left(\frac{L}{N}\right)^2
\]
If $\frac{L}{N}=c$, $b$ is not changed by $N$. To test this counter-intuitive inference, we let $L=N=$ \{100,\ 200,\ 500,\ 800,\ 1000,\ 2000,\ 5000\}. Since $\frac{L}{N}=1$, $b$ and $E[\alpha]$ are unchanged. But Table~\ref{tab:1} shows the real training accuracies vary with $N$. The predicted training accuracy is close to the real training accuracy only at $N=200$ and the real training accuracy decreases with the growth of $N$. Hence, our theory must be refined using empirical corrections.

\begin{table}[h]
\caption{
Accuracy results comparison. The columns from left to right are dimension, dataset size, number of neurons in hidden layer, the real training accuracy and estimated training accuracy by Eq.~(\ref{eq:14}) and Theorem \ref{thm:1}.
}
\centering
\begin{tabular}{lllll}
\hline\noalign{\smallskip}
$d$ & $N$ & $L$ & Real Acc & Est. Acc  \\
\noalign{\smallskip}\hline\noalign{\smallskip}
2 & 100 & 100 & 0.844 & 0.769 \\
2 & 200 & 200 & 0.741 & 0.769 \\
2 & 500 & 500 & 0.686 & 0.769 \\
2 & 800 & 800 & 0.664 & 0.769 \\
2 & 1000 & 1000 & 0.645 & 0.769 \\
2 & 2000 & 2000 & 0.592 & 0.769 \\
2 & 5000 & 5000 & 0.556 & 0.769 \\
\noalign{\smallskip}\hline
\end{tabular}
\label{tab:1}
\end{table}

The correction could be applied on either Eq.~(\ref{eq:14}) or (\ref{eq:11}). We modify Eq.~(\ref{eq:14}) because the range of function~(\ref{eq:11}) is $(0.5, \ 1)$, which is an advantage for training accuracy estimation. In Table~\ref{tab:1}, the real training accuracy decreases when $N$ increases; this suggests that the exponent of $N$ in Eq.~(\ref{eq:14}) should be larger than that of $L$. Therefore, according to (\ref{eq:14}), we consider a more general equation to connect the ensemble index $b$ with three \textbf{spatial parameters} $x_d$, $y_d$, and $c_d$:
\begin{equation} \label{eq:15}
\boxed{
b=c_d\frac{L^{x_d}}{N^{y_d}}
}
\end{equation}

\begin{obs} \label{obs:1}
The ensemble index $b$ is computed by Eq.~(\ref{eq:15}) with spatial parameters $\left\{x_d,\ y_d,\ c_d\right\}$, where $x_d,\ y_d$ are exponents of $N$ and $L$, and $c_d$ is a constant. All the spatial parameters vary with the dimensionality of inputs $d$.
\end{obs}

In 2-D, to determine the $x_2,\ y_2,\ c_2$ in Eq.~(\ref{eq:15}), we test 81 $\left\{N,\ L\right\}$, which are the combinations of: $L,\ N\ \in$ \{100, 200, 500, 800, 1000, 2000, 5000, 10000, 20000\}. For each $\left\{N[i],\ L\left[i\right]\right\}$, we could obtain a real training accuracy by experiment. Their corresponding ensemble indexes $b[i]$ are found using Eq.~(\ref{eq:11}). Finally, we determine the $x_2,\ y_2,\ c_2$ by fitting the $\frac{1}{b},\ N,\ L$ to Eq.~(\ref{eq:15});  this yields the expression for the ensemble index for 2-D:
\begin{equation} \label{eq:16}
b=8.4531\frac{L^{0.0744}}{N^{0.6017}}
\end{equation}

\begin{figure}
    \centering
    \includegraphics[width=0.45\textwidth]{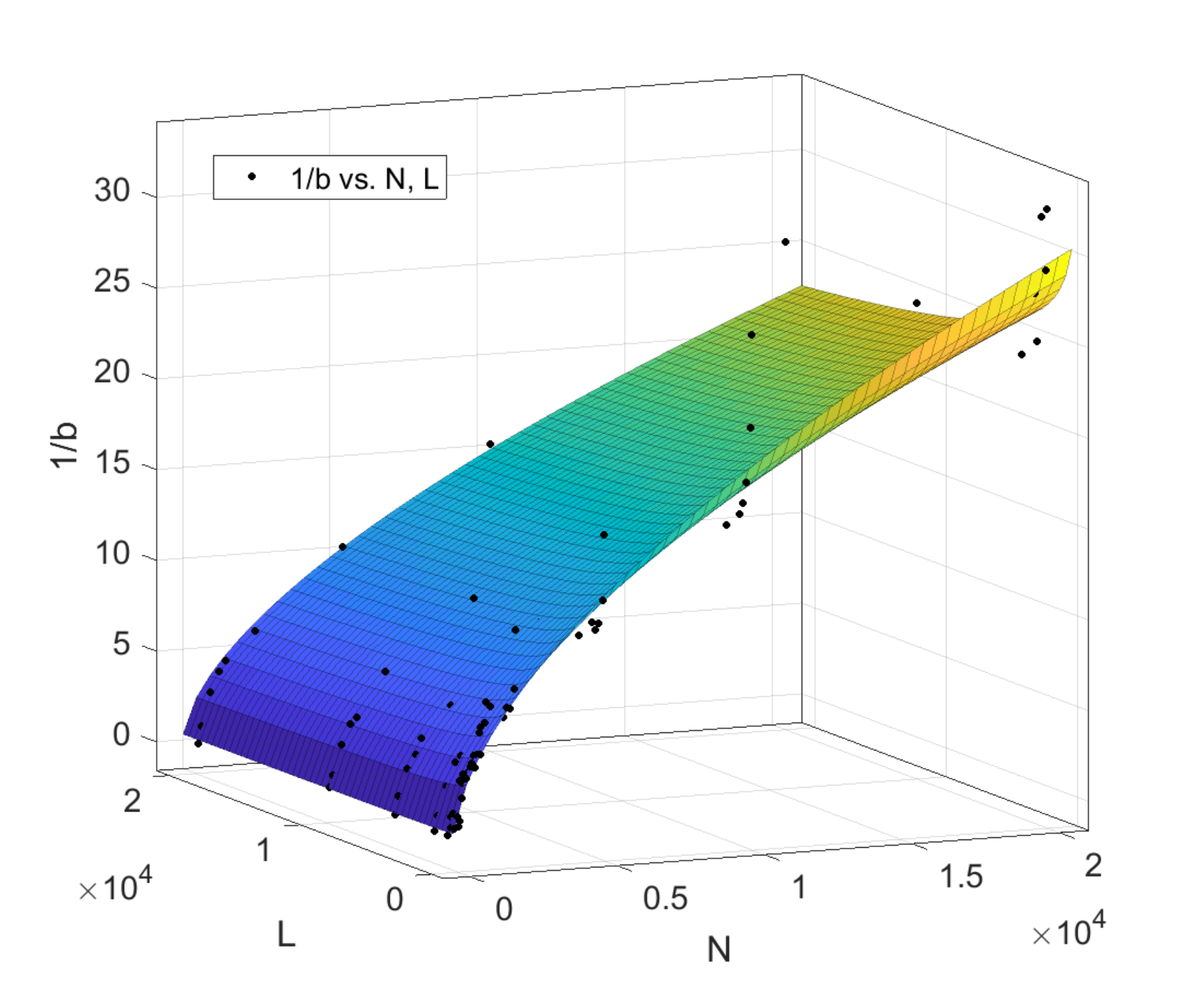}
    \caption{
    Fitting curve of $\frac{1}{b}=f(N,L)$ in 2-D
    }\label{fig:2}
\end{figure}

\begin{table}[h]
\caption{
Estimated training accuracy results comparison in 2-D. The columns from left to right are dataset size, number of neurons in hidden layer, the real training accuracy, estimated/predicted training accuracy by Eq.~(\ref{eq:16}) and Theorem~\ref{thm:1}, and (absolute) differences based on estimations between real and estimated accuracies.
}
\centering
\begin{tabular}{llllll}
\hline\noalign{\smallskip}
$d=2$ & $N$ & $L$ & Real Acc & Est. Acc & Diff \\
\noalign{\smallskip}\hline\noalign{\smallskip}
 & 115 & 2483 & 0.910 & 0.846 & 0.064 \\
 & 154 & 1595 & 0.805 & 0.819 & 0.014 \\
 & 243 & 519 & 0.782 & 0.767 & 0.015 \\
 & 508 & 4992 & 0.699 & 0.724 & 0.025 \\
 & 689 & 2206 & 0.665 & 0.685 & 0.020 \\
 & 1366 & 4133 & 0.614 & 0.631 & 0.016 \\
 & 2139 & 2384 & 0.578 & 0.593 & 0.015 \\
 & 2661 & 890 & 0.566 & 0.573 & 0.007 \\
 & 1462 & 94 & 0.577 & 0.592 & 0.014 \\
 & 3681 & 1300 & 0.555 & 0.560 & 0.004 \\
 & 4416 & 4984 & 0.556 & 0.559 & 0.003 \\
 & 4498 & 1359 & 0.550 & 0.552 & 0.002 \\
\noalign{\smallskip}\hline
\end{tabular}
\label{tab:2}
\end{table}

The fitting process uses the Curve Fitting Tool (cftool) in MATLAB. Figure~\ref{fig:2} shows the 81 points of $\left\{N,\ L\right\}$ and the fitted curve. The R-squared value of fitting is about 0.998. The reason to fit $\frac{1}{b}$ instead of $b$ is to avoid $b=+\infty$ when the real accuracy is 1 (which can occur). In this case, $\frac{1}{b}=0$. Conversely, $b=0$ when the real accuracy is 0.5, which never appears in our experiments. Using an effective classifier rather than random guess makes the accuracy $>0.5$ ($b>0$), thus, $\frac{1}{b}\neq+\infty$. To cover the parameter space as completely as possible, we manually choose the 81 points of $N$ and $L$. We then verify the fitted model (Eq.~\ref{eq:16}) by other random values of $N$ and $L$.

\begin{table}[h]
\caption{
Spatial parameters $\left\{x_d,\ y_d,\ c_d\right\}$ in Eq.~(\ref{eq:15}) (Observation~\ref{obs:1}) for various dimensionalities of inputs are determined by fitting.
}
\centering
\begin{tabular}{lllll}
\hline\noalign{\smallskip}
$d$ & $x_d$ & $y_d$ & $c_d$ & R-squared \\
\noalign{\smallskip}\hline\noalign{\smallskip}
2 & 0.0744 & 0.6017 & 8.4531 & 0.998 \\
3 & 0.1269 & 0.6352 & 15.5690 & 0.965 \\
4 & 0.2802 & 0.7811 & 47.3261 & 0.961 \\
5 & 0.5326 & 0.8515 & 28.4495 & 0.996 \\
6 & 0.4130 & 0.8686 & 61.0874 & 0.996 \\
7 & 0.4348 & 0.8239 & 33.4448 & 0.977 \\
8 & 0.5278 & 0.9228 & 61.3121 & 0.996 \\
9 & 0.7250 & 1.0310 & 82.5083 & 0.995 \\
10 & 0.6633 & 1.0160 & 91.4913 & 0.995 \\
\noalign{\smallskip}\hline
\end{tabular}
\label{tab:3}  
\end{table}

By using Eqs.~(\ref{eq:16}) and (\ref{eq:11}), we estimate training accuracy on random values of $N$ and $L$ in 2-D. The results are shown in Table~\ref{tab:2}. The differences between real and estimated training accuracies are small, except the first (row) one. For higher real-accuracy cases ($>~0.86$), the difference is larger because $\frac{1}{b}<1$ ($b>1$ when the accuracy $> 0.86$), while the effect is smaller in cases with $\frac{1}{b}>1$ during the fitting to find Eq.~(\ref{eq:16}).

\subsection{Three and more dimensions; empirical correction procedure}
\label{sec:3.2}
We repeat the same processes as for 2-D to determine spatial parameters $\left\{x_d,\ y_d,\ c_d\right\}$ in Observation~\ref{obs:1} for data dimensionality from 3 to 10. Results are shown in Table~\ref{tab:3}. The R-squared values of fitting are high.

\begin{figure}
    \centering
    \includegraphics[width=0.45\textwidth]{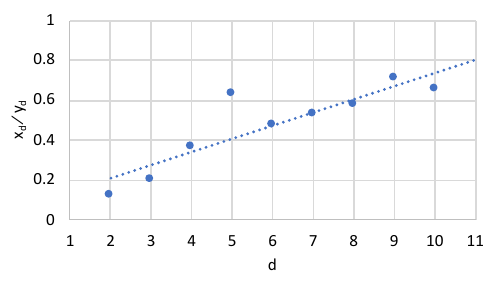}
    \caption{
    Plots of $d\ \mbox{v.s.}\ \frac{x_d}{y_d}$ from Table~\ref{tab:3}. Blue dot-line is linearly fitted by points to show the growth.
    } \label{fig:3}
\end{figure}

Such results reaffirm the necessity of correction in our theory because, when compared to Eq.~(\ref{eq:14}), spatial parameters $\left\{x_d,\ y_d,\ c_d\right\}$ are not $\left\{d,\ 2,\ \frac{1}{d!}\right\}$. But the growth of $\frac{x_d}{y_d}$ is preserved. From Eq.~(\ref{eq:14}),
\[
\frac{x_d}{y_d}=\frac{d}{2}
\]
The $\frac{x_d}{y_d}$ linearly increases with $d$. The real $d\ \mbox{v.s.} \ \frac{x_d}{y_d}$ (Figure~\ref{fig:3}) shows the same tendency.

Table~\ref{tab:3} indicates that $x_d,\ y_d, and\ c_d$ increase almost linearly with $d$. Thus, we apply linear fitting on $d\mbox{-}x_d$, $d\mbox{-}y_d$, and $d\mbox{-}c_d$ to obtain these fits:

\begin{equation} \label{eq:17}
\boxed{
\begin{array}{l p{1em} l}
x_d=0.0758\cdot d - 0.0349 & & (R^2 = 0.858)\\
y_d=0.0517\cdot d + 0.5268 & & (R^2 = 0.902)\\
c_d=9.4323\cdot d - 8.8558 & & (R^2 = 0.804)
\end{array}
}
\end{equation}

Eq.~(\ref{eq:17}) is a supplement to Observation~\ref{obs:1}, which employs empirical corrections to Eq.~(\ref{eq:14}) for determining the ensemble index $b$ of Theorem~\ref{thm:1}.

\subsection{Summary of the algorithm}
\label{sec:sum_alg}
To associate the two statements of Observation~\ref{obs:1} and Theorem~\ref{thm:1}, we can estimate the training accuracy for a two-layer FCNN on two-class random datasets using only three arguments: the dimensionality of inputs ($d$), the number of data points ($N$), and the number of neurons in the hidden layer ($L$), without actual training.

\begin{algorithm}[h]
    \caption{To estimate the training accuracy $\alpha$ for two-layer neural networks on a random dataset without training}\label{algorithm}
    
    \KwIn{the dimensionality of inputs $d$, the number of data points $N$, the number of neurons in the hidden layer $L$.}
    \KwResult{the expectation of training accuracy $E\left[\alpha\right]$.}
    
    \tcp{Use Eq.~(\ref{eq:17}) to calculate the three spatial parameters: $\left\{x_d,\ y_d,\ c_d\right\}$.}
    $x_d\leftarrow 0.0758\cdot d - 0.0349$\;
    $y_d\leftarrow 0.0517\cdot d + 0.5268$\;
    $c_d\leftarrow 9.4323\cdot d - 8.8558$\;
    
    \tcp{Use Eq.~(\ref{eq:15}) to compute the ensemble index $b$.}
    $b\leftarrow c_d\frac{L^{x_d}}{N^{y_d}}$\;
    
    \tcp{Use Eq.~(\ref{eq:11}) to compute the expectation of training accuracy $E\left[\alpha\right]$.}
    $E\left[\alpha\right]\leftarrow \frac{1}{2}+\frac{\sqrt{2\pi b}}{4}\mbox{erfi} {\left(\frac{1}{\sqrt{2b}}\right)}e^{-\left(\frac{1}{2b}\right)}$\;
    
    \KwOut{$E\left[\alpha\right]$.}
\end{algorithm}

\subsection{Testing}
\label{sec:3.3}
\begin{figure}
    \includegraphics[width=0.45\textwidth]{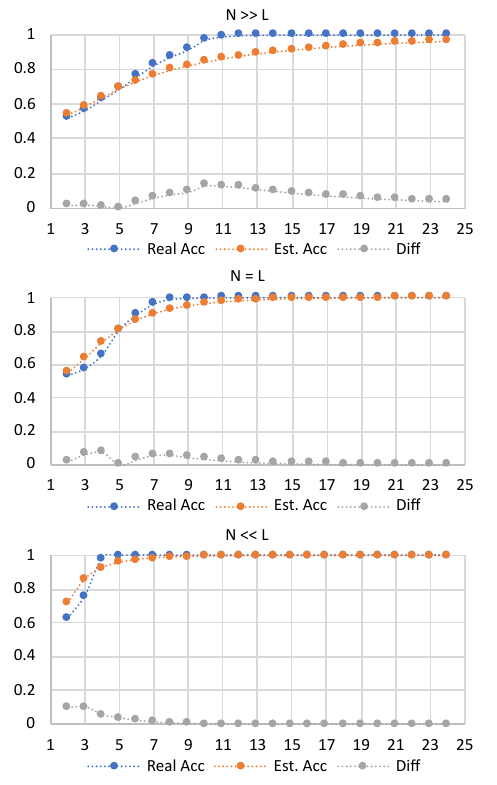}
    \caption{
    Estimated training accuracy results comparisons. y-axis is accuracy, x-axis is the dimensionality of inputs ($d$).
    } \label{fig:4}
\end{figure}

To verify our theorem and observation, we first estimate training accuracy on a larger range of dimensions ($d=2$ to 24) and for three situations:
\begin{itemize}
  \item $N \gg L: N=10000,\ L=1000$
  \item $N\cong L: N=L=10000$
  \item $N\ll L: N=1000,\ L=10000$
\end{itemize}

The results are shown in Figure~\ref{fig:4}. The maximum differences between real and estimated training accuracies are about 0.130 ($N\gg L$), 0.076 ($N\cong L$), and 0.104 ($N\ll L$). There may be two reasons that the differences are not small in some cases of $N\gg L$: 1) in the fitting process, we do not have enough samples for which $N\gg L$, so that the corrections are not perfect; 2) the reason why the differences are greater for higher real accuracies is similar to the 2-D situation discussed above.

\begin{figure}
    \includegraphics[width=0.45\textwidth]{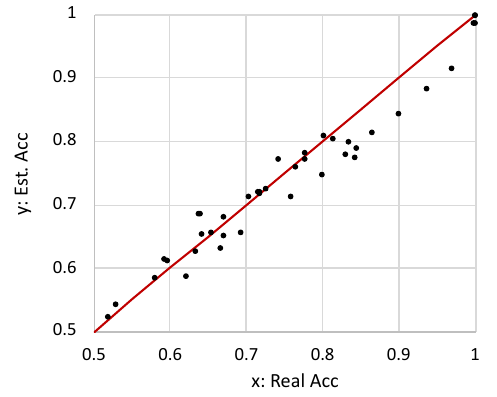}
    \caption{
    Evaluation of estimated training accuracy results. y-axis is estimated accuracy; x-axis is the real accuracy; each dot is for one case; red line is $y=x$. R-squared $\approx0.955$.
    } \label{fig:5}
\end{figure}

In addition, we estimate training accuracy on 40 random cases. For each case, the $N,\ L\in [100,\ 20000]$ and $d\in [2,\ 24]$, but principally we use $d\in [2,\ 10]$ because in high dimensions, almost all cases’ accuracies are close to 100\% (see Figure~\ref{fig:4}). Figure~\ref{fig:5} shows the results. Each case is plotted with its real and estimated training accuracy. The overall R-squared value is about 0.955, indicating good estimation.

\section{Discussion}
\label{sec:4}

\subsection{Significance and contributions}
\label{sec:4.1}

Our main contribution is to build a novel theory to estimate the training accuracy for a two-layer FCNN used on two-class random datasets without using input data or trained models (training); the theory could help to understand the mechanisms of neural network models. The estimation uses three arguments:
\begin{enumerate}
    \item Dimensionality of inputs ($d$)
    \item Number of data points ($N$)
    \item Number of neurons in the hidden layer ($L$)
\end{enumerate}
It also is based on the conditions of classifier models and datasets stated in the Sec.~\ref{sec:prelim} and the two hypotheses. There appear to be no other studies that have proposed a method to estimate training accuracy in this way.

Our theory is based on the notion that hidden layers in neural networks perform space partitioning and the hypothesis that the data separation ratio determines the training accuracy. Theorem~\ref{thm:1} introduces a mapping function between training accuracy and the ensemble index. The ensemble index, by virtue of its domain $(0,\ \infty)$; is better suited than accuracy (domain $[0.5,\ 1]$) for the computation required by the fitting process. This is to maintain parity between the input variables’ domain and the range of the fitted quantity (the ensemble index). The extended domain consists with the domains of $N$, $L$, and $d$; it is good for designing prediction models or fitting experimental data. Observation~\ref{obs:1} provides a calculation of the ensemble index based on empirical corrections. And these corrections; which successfully improve our model to make estimations.

\subsection{Limitations and future work}
\label{sec:4.2}

\subsubsection{Improvements of the theory}
\label{sec:4.2.1}
An ideal theory is to precisely estimate the accuracy with no, or very limited, empirical corrections. Our purely theoretical results (to estimate training accuracy by Eq.~(\ref{eq:14}) and Theorem~\ref{thm:1}) cannot match the real accuracies in 2-D (Table~\ref{tab:1}). Empirical corrections are therefore required.

Those corrections are not perfect so far, because of the limited number of fitting and testing samples. Although the empirically-corrected estimation is not very accurate for some cases, the method does reflect some characteristics of the real training accuracies. For example, the training accuracies of the $N\gg L$ cases are smaller than those of $N\ll L$, and for specific $N$ and $L$, the training accuracies of higher dimensionality of inputs are greater than those of lower dimensionality. These characteristics are shown by the estimation curves in Figure~\ref{fig:4}. Although there are large errors for some cases, Figure~\ref{fig:4} shows the similar tendencies of real and estimated accuracy.

And the theorem, Observation~\ref{obs:1} and empirical corrections could be improved in the future. The improvements would be along these directions:
\begin{enumerate}
    \item To use more data for fitting.
    \item To rethink the space partitioning problem to change Observation~\ref{obs:1}. We could use a different approximation formula from Eq.~(\ref{eq:13}), or involve the probability of reaching the maximum number of partitions.
    \item To modify Theorem~\ref{thm:1} by reconsidering the necessity of complete separation. In fact, in real classification problems, \textbf{complete separation of all data points is too strong a requirement}. Instead, to require only that no different-class samples are assigned to the same partition would be more appropriate.
    \item To involve the capacity of a separating plane \cite{duda_pattern_2012}:
    \[
    f(N,d)=\left\{
    \begin{array}{l p{1em} c}
    1& & N\le d+1 \\
    \displaystyle \frac{2}{2^N}\sum_{i=0}^{d}{N-1 \choose i} & & N>d+1
    \end{array}\right.
    \]
    Where we would estimate $f(N,d)$, the probability of the existence of a hyper-plane separating two-class $N$ points in $d$ dimensions.
\end{enumerate}

\subsubsection{For deeper neural networks}
\label{sec:4.2.2}
Our proposed estimation theory could extend to multi-layer neural networks. As discussed at the beginning of the second section, $L$ neurons in the first hidden layer assign every input a unique code. The first hidden layer transforms $N$ inputs from $d$-D to $L$-D. Usually, $L>d$, and the higher dimensionality makes \textbf{data separation easier for successive layers}. Also, the effective $N$ decreases when data pass through layers if we consider \textbf{partition merging}. Specifically, if a partition and its neighboring partitions contain same-class data, they could be merged into one because these data are locally classified well. The decrease of actual inputs makes data separation easier for successive layers also. 

Alternatively, the study of Pascanu et al.~\cite{pascanu_number_2014} provides a calculation of the number of space partitions ($S$ regions) created by $k$ hidden layers:
\[
S=\left(\prod_{i=1}^{k-1}\left\lfloor\frac{n_i}{n_0}\right\rfloor\right)\sum_{i=0}^{n_0}{n_k \choose i}
\]
Where, $n_0$ is the size of input layer and $n_i$ is the $i$-th hidden layer. This reduces to Eq.~(\ref{eq:12}) in the present case ($k=1$, $n_0=d$ and $n_1=L$). The theory for multi-layer neural networks could begin by using the approaches above.

\subsubsection{More conditions}
\label{sec:4.2.3}
In addition, this study still has several places that are worth attempting to enhance or extend and raises some questions for future work. For example, the proposed theory could also extend to distributions of data other than uniform, unequal number of samples in different classes, and/or other types of neural networks by modifying the ways to calculate separation probabilities, such as in Eqs.~(\ref{eq:1}) and (\ref{eq:6}).  

\section{Conclusions}
\label{sec:5}
In this study, we propose a novel theory based on space partitioning to estimate the approximate training accuracy for two-layer neural networks on random datasets without training. It does this using only three arguments: the dimensionality of inputs ($d$), the number of input data points ($N$), and the number of neurons in the hidden layer ($L$). The theory has been verified by the computation of real training accuracies in our experiments. Although the method requires empirical correction factors, they are determined using a principled and repeatable approach. The results indicate that the method will work for any dimension, and it has the potential to estimate deeper neural network models. It is not the purpose of this work, however, to examine generalizability to arbitrary datasets. This study may raise other questions and suggest a starting point for a new way for researchers to make progress on studying the \textit{transparency} of deep learning.

\section*{Appendix: Simplification from Eq.~(\ref{eq:3}) to Eq.~(\ref{eq:4})}
\label{sec:appx}
\addcontentsline{toc}{section}{Appendix}

In the paper, Eq.~(\ref{eq:3}):
\[\lim_{N \to +\infty}P_c=\lim_{N \to + \infty}\left(\frac{1}{e}\right)^N{\left(\frac{bN^a}{bN^a-N}\right)^{bN^a-N+0.5}}\]
Ignoring the constant 0.5 (small to $N$),
\[\lim_{N \to +\infty}P_c=\lim_{N\to+\infty}\left(\frac{1}{e}\right)^N{\left(\frac{bN^a}{bN^a-N}\right)^{bN^a-N}}\]
Using equation ${x=e}^{\ln{x}}$,
\begin{multline*}
    \lim_{N \to +\infty}P_c={\lim_{N \to +\infty}e}^{\ln{\left[\left(\frac{1}{e}\right)^N{\left(\frac{bN^a}{bN^a-N}\right)^{bN^a-N}}\right]}} \\
    ={\lim_{N \to +\infty}e}^{{
    \underbrace{-N+\left(bN^a-N\right)\ln{\left(\frac{bN^a}{bN^a-N}\right)}}_{(\mathcal{A})}}} 
\end{multline*}
Let $t=\frac{1}{N}\to +0$,
\begin{multline*}
\left(\mathcal{A}\right)=-\frac{1}{t}+\left(\frac{b}{t^a}-\frac{1}{t}\right)\ln{\left(\frac{\frac{b}{t^a}}{\frac{b}{t^a}-\frac{1}{t}}\right)}\\ =\frac{\left(b-t^{a-1}\right)\ln\left(\frac{b}{b-t^{a-1}}\right)-t^{a-1}}{t^a}
\end{multline*}
[i] If $a=1$,
\[(\mathcal{A})=\frac{\overbrace{\left(b-1\right)\ln\left(\frac{b}{b-1}\right)}^{(\mathcal{B})}-1}{t}\]
\[\left(\mathcal{B}\right)=\ln\left(\frac{b}{b-1}\right)^{\left(b-1\right)}
\]
In $R$, it is easy to show that, for $b>0$,
\[
1<\left(\frac{b}{b-1}\right)^{\left(b-1\right)}<e\]
Then,
\[0<\left(\mathcal{B}\right)<1
\]
\[\lim_{t\to+0}\left(\mathcal{A}\right)=\lim_{t\to+0}\frac{\left(\mathcal{B}\right)-1}{t}=-\infty
\]
Therefore,
\[\lim_{N\to+\infty}P_c=e^{-\infty}=0 \tag{Eq.~(\ref{eq:4}) in paper, when $a=1$}
\]
[ii] For $a>1$, by applying L'H\^{o}pital's rule several times:
\begin{multline*}
\displaystyle \lim_{t\to+0}\left(\mathcal{A}\right)=\lim_{t\to+0}\frac{\left(b-t^{a-1}\right)\ln\left(\frac{b}{b-t^{a-1}}\right)-t^{a-1}}{t^a} \\
\displaystyle \myeq\lim_{t\to+0}\frac{(1-a)t^{a-2}\ln\left(\frac{b}{b-t^{a-1}}\right)}{at^{a-1}} \\ =\lim_{t\to+0}\frac{(1-a)\ln\left(\frac{b}{b-t^{a-1}}\right)}{at} \\
\displaystyle \myeq\lim_{t\to+0}\frac{\left(a-1\right)^2}{a\left(t-bt^{2-a}\right)}=\lim_{t\to+0}\frac{\frac{\left(a-1\right)^2}{t}}{a\left(1-bt^{1-a}\right)} \\
\displaystyle \myeq\lim_{t\to+0}\frac{\frac{\left(a-1\right)^2}{-t^2}}{\left(a-1\right)abt^{-a}}=\lim_{t\to+0}-\frac{\left(a-1\right)}{abt^{2-a}}
\end{multline*}
Substitute $N=\frac{1}{t}$,
\[
\lim_{N\to+\infty}P_c=\lim_{t\to+0}e^{\left(\mathcal{A}\right)}=\lim_{N\to+\infty}e^{-\frac{\left(a-1\right)N^{2-a}}{ab}} \tag{Eq.~(\ref{eq:4}) in paper, when $a>1$}
\]
When $1<a<2$ (and $a=1$, shown before in [i]),
\[\lim_{N\to+\infty}P_c=e^{-\frac{+\infty}{ab}}=0\]
When $a>2$,
\[\lim_{N\to+\infty}P_c=e^{-\frac{0}{ab}}=1\]
For $a=2$,
\[\lim_{N\to+\infty}P_c=e^{-\left(\frac{1}{2b}\right)} \tag{Eq.~(\ref{eq:5}) in paper}
\]
$\blacksquare$


%

\bibliographystyle{spmpsci}      
\bibliography{ref}   


\end{document}